# Gradient Domain Weighted Guided Image Filtering

Bo Wang[a,b], Yihong Wang[a,b], Xiubao Sui[a,b,*], Yuan Liu[a,b], Qian Chen[a,b]

[a] *School of Electronic and Optical Engineering, Nanjing University of Science and Technology, Nanjing 210094, China*

[b] *Jiangsu Key Laboratory of Spectral lmaging and Intelligent Sense, Nanjing University of Science and Technology, Nanjing, Jiangsu Province 210094,China*

[*]*Corresponding author at: Nanjing University of Science and Technology, Nanjing, Jiangsu Province 210094,China*

E-mail address: sxb@njust.edu.cn (Xiubao Sui).

**Abstract**

Guided image filter is a well-known local filter in image processing. However, the presence of halo artifacts is a common issue associated with this type of filter. This paper proposes an algorithm that utilizes gradient information to accurately identify the edges of an image. Furthermore, the algorithm uses weighted information to distinguish flat areas from edge areas, resulting in sharper edges and reduced blur in flat areas. This approach mitigates the excessive blurring near edges that often leads to halo artifacts. Experimental results demonstrate that the proposed algorithm significantly suppresses halo artifacts at the edges, making it highly effective for both image denoising and detail enhancement.

**Keywords**

Edge-preserving filtering、Average strategy、detail enhancement、image denoising、halo artifact

1. **INTRODUCTION**

Edge preservation is a widely employed technique in various fields, including image processing, computer vision, and camera measurement. One such application of edge preservation is image smoothing [1], which typically results in the blurring of edges. However, if an image can be smoothed while preserving its edges, the resulting image quality can be significantly improved. In single-frame image enhancement [2], the process involves subtracting the input from the output to obtain a detail layer, which is then linearly enlarged and superimposed onto the input. The inclusion of edge information in the detail layer further enhances the imaging quality. Thus, edge-preserving smoothing algorithms can be utilized for image detail enhancement. Furthermore, image denoising often results in the loss of edge information, which can be mitigated by applying smoothing algorithms with edge preservation properties.

A variety of edge-preserving image denoising algorithms can be broadly categorized into two categories: global optimization algorithms and local filtering algorithms. The former category includes algorithms such as total variation (TV) [4][12], weighted least squares [3][13], and L0 parametric gradient minimization method [2]. These algorithms combine different regularization and fidelity terms into an optimization problem that is solved through multiple iterations. However, global edge-preserving algorithms typically suffer from slow runtime, excessive memory consumption, and gradient inversion. The latter category of algorithms includes faster and more memory-efficient methods such as median filter [31][32], bilateral filtering [5][14], and guided image filtering (GIF) [6]. While bilateral filtering is a widely used weighted mean filter with fast operation, it suffers from gradient inversion [15][16] [29] when applied to image detail enhancement, which degrades the image quality. In contrast, the guided filter (GIF) [6] is a linear filter that can avoid gradient inversion. However, GIF tends to produce halo artifacts in the edge region. To address this problem, researchers have proposed a series of improved algorithms. For instance, Li et al. proposed the weighted guided image filter (WGIF) [7], which introduces an edge-aware weighting operator in the guided filter to improve its edge protection property. Inspired by the Gradientshop [17][18], Kou et al. proposed the gradient-domain guided image filter (GDGIF) [8], which combines the first-order edge-aware constraint to further improve the edge protection property of the guided filter and mitigate the halo artifacts at the edges of the guided filter. Xie et al. proposed an improved guided filter algorithm incorporating gradient information [9], which can adaptively distinguish and emphasize the edges using exponential function framework to design weights that control the smoothing multiplier in different image regions. Chen proposed a weighted aggregated guided filtering (WAGIF) [10] that can obtain sharp edges and avoid halo artifacts using a mean-value strategy for handling overlapping windows. Sun et al. proposed guided filtering with steering kernel (SKWGIF) [11], which can adaptively learn the directional information of edges and obtain high-quality images while mitigating edge artifacts. However, SKWGIF [11] is not suitable for engineering practice due to its high memory consumption and long running time. While the edge-aware operator proposed by WGIF [7] and Xie [9] lacks explicit edge constraints and is prone to edge blurring, GDGIF [8] has limited protection for tiny edges, and WAGIF [10] only improves the guided filtering weighting strategy without explicit edge protection.

This paper proposes a novel gradient-weighted guided filtering algorithm to address the limitations of existing guided image filter and its derived versions, namely GIF、WGIF、GDGIF 、Xie and WAGIF. The proposed algorithm combines gradient information and weighted information and introduces edge-aware and edge-preserving operators, as well as pixel weight distribution, to address halo artifacts, excessive image smoothing, and enhance image edge preservation. Unlike GDGIF, the edge-aware operator in the proposed algorithm utilizes gradient and edge change information to accurately distinguish edge regions from flat regions. Furthermore, the proposed edge protection operator uses fast gradient calculation to protect the edge pixels while maintaining computation speed. Additionally, a new overlap window computation strategy is introduced in this algorithm to improve its edge-preserving property, which uses accurate image weight distribution. Experimental evaluations of the proposed algorithm for image detail enhancement and image denoising show that it outperforms existing methods, including GIF, WGIF, GDGIF, Xie, and WAGIF, demonstrating the efficacy of the proposed approach.

The rest of this paper is organized as follows. GIF, edge-preserving and smoothing properties are summarized in Section II. Section III describes the details of GWGIF. The GWGIF is analyzed and verified in Section IV. Two applications of GWGIF are given in Section V. Section VI provides concluding remarks.

## 2. RELATED WORKS ON GUIDED IMAGE FILTERING

### 1) Guided Image Filtering

The guided filter is closely related to the matting Laplacian matrix [30].The initial model of GIF is a linear model with input data consisting of an input image X and a guide image G. Assume that the output image $Z$ is a phenomenal transformation of the guide image $G$ in the window $\Omega_{\xi 1}$[19],[20]:

$$Z_{(p)} = a_{p'}G_{(p)} + b_{p'}, \quad \forall p \in \Omega_{\xi 1}(p') \tag{1}$$

where $Z_{(p)}$ denotes the output image. $G_{(p)}$ denotes the input guide image. $p'$ represents the current pixel and p represents the pixel in the window corresponding to the current pixel point $p'$. $a_{p'}$ and $b_{p'}$ sourced from linear ridge regression model:

$$E = \sum_{p \epsilon \Omega_{\xi 1}(p')} [(a_{p'}G_{(p)} + b_{p'} - X(p))^2 + \lambda a_{p'}^2] \tag{2}$$

the objective is to find the value that minimizes the gap between the output image and the desired filtered image while satisfying the linear model $a_{p'}$ and $b_{p'}$, $\lambda$ is the regularization parameter of the penalty term $a_{p'}$. The final values of $a_{p'}$ and $b_{p'}$ are as follows:

$$a_{p'} = \frac{\mu_{G \odot X, \xi 1}(p') - \mu_{G,\xi 1}(p')\mu_{X,\xi 1}(p')}{\sigma_{G,\xi 1}^2(p') + \lambda} \tag{3}$$

$$b_{p'} = \mu_{X,\xi 1}(p') - a_{p'}\mu_{G,\xi 1}(p') \tag{4}$$

where $\odot$ denotes the multiplication of two matrix elements. $\xi 1$ denotes the size of the window radius. $\mu_{G \odot X, \xi 1}(p')$、$\mu_{G,\xi 1}(p')$ and $\mu_{X,\xi 1}(p')$ denotes the means of the matrix $GX$、$G$ and $X$ in the window $\Omega_{\xi 1}(p')$. $\sigma_{G,\xi 1}^2(p')$ represents the variance of the matrix $G$ in the window $\Omega_{\xi 1}(p')$.

Since the same pixel is counted by several different $\Omega_{\xi 1}(p')$ windows at the same time and the $a_{p'}$ and $b_{p'}$ are computed in different windows will produce different values. Therefore, the values within the overlap window need to be normalized. GIF uses the mean value strategy: average over all a and b. The final output is

$$Z_{(p)} = \overline{a_{p'}} G_{(p)} + \overline{b_{p'}} \tag{5}$$

where $\overline{a_{p'}} = \frac{1}{|\Omega_{\xi_1}(p)|} \sum_{p \in \Omega_{\xi_1}(p')} a_{p'}$, $\overline{b_{p'}} = \frac{1}{|\Omega_{\xi_1}(p)|} \sum_{p \in \Omega_{\xi_1}(p')} b_{p'}$, $|\Omega_{\xi_1}(p)|$ is the normalization factor of $\Omega_{\xi_1}(p)$ to avoid data overflow.

### 2) Edge Preservation Analysis

Analyzing equation (1), it is easy to see that the value of $a_{p'}$ determines the pixel distribution at the edge of the image and the value of $b_{p'}$ determines the pixel distribution in the flat region. Calculate the gradients on both sides of equation (1).

$$\nabla Z_{(p)} = a_{p'} \nabla G_{(p)} \tag{6}$$

The output $Z_{(p)}$ contains less edge information in the guide image when $a_{p'}$ is smaller, the output $Z_{(p)}$ contains more edge information in the guide image when $a_{p'}$ become larger. Therefore, the value of a determines the edge-preserving nature of the GIF. The major method for eliminating edge halo artifacts is to provide a more accurate description of the edge information of the image.

Further analysis of equation (3): when the pixel is located in the edge region, the image contains more scene information, the image gradient is large and the center pixel within the filter window has a large dispersion from the surrounding pixels. Therefore the image variance is large: $\sigma_{G,\xi_1}^2(p') >> \lambda$, $a_{p'} \approx 1$; When the pixels are located in a flat area, the image contains less scene information, the image gradient is small and the central pixel within the filter window is less discrete from the surrounding pixels. Thus, the image variance is small: $\sigma_{G,\xi_1}^2(p') << \lambda$, $a_{p'} \approx 0$. In summary: GIF has edge-preserving properties. However, due to the fixed value of the regularization parameter $\lambda$, the consistent $\lambda$ norm intensity tends to cause some edges to be over-smoothed and produce halo artifacts when dealing with edges in different areas. An adaptive guided filtering has been proposed in [7] to solve this problem by replacing equation (2) with the following equation.

$$E = \sum_{p \epsilon \Omega_{\xi_1}(p')} [(a_{p'}G_{(p)} + b_{p'} - X(p))^2 + \frac{\lambda}{\Gamma_G(p')} a_{p'}^2] \quad (7)$$

where $\Gamma_G(p')$ is defined as an edge-aware constraint consisting of the variance of the local 3*3 window domain.

$$\Gamma_G(p') = \frac{1}{N} \sum_{p=1}^{N} \frac{\sigma_{G,1}^2(p') + \varepsilon}{\sigma_{G,1}^2(p) + \varepsilon} \quad (8)$$

$\sigma_{G,1}^2(p')$ is the variance within the window $\Omega_{\xi_1}(p)$. $\varepsilon$ is a smaller positive constant defined as $(0.0001 * L)^2$. L is defined as the dynamic range of the input image. The role of $\Gamma_G(p')$ is to constrain the regularization parameter $\lambda$ in Eq. 3, which in turn affects $a_{p'}$ for the purpose of edge constraint. The algorithm proposed in this paper uses a new adaptive edge constraint.

### 3) Smoothing Property Analysis

Image smoothing[22] is the core step of image denoising, such as mean filtering[23] and median filtering[24], both of which assign values to the central pixels within the window. Thus, the effect of noise is attenuated. Analytical equation (5): since each pixel needs to be computed in one of the many overlapping windows, it is necessary to normalize the pixels within the different windows. GIF adopts an averaging strategy (mean filter): $a_{p'}$ and $b_{p'}$ are averaged within different windows. Averaging strategy is a common method to handle overlapping windows. However, due to the homogeneity of the mean filter, the mean filter assigns the same weight to all pixels within its local window, ignoring the fact that there are differences between pixels in the input image. This leads to excessive smoothing of pixels in flat regions and loss of detail information, which affects the image quality. Therefore, it is necessary to change the averaging strategy of GIF to avoid the excessive smoothing and improve the image quality.

## 3. GRADIENT DOMAIN WEIGHTED GUIDED FILTERING

In this paper, we propose a gradient-domain guided filtering combined with weight distribution, including edge-aware constraint, explicit first-order edge protection constraint and weighted-mean strategy.

The flow chart of the algorithm in this paper is shown in Fig.1.

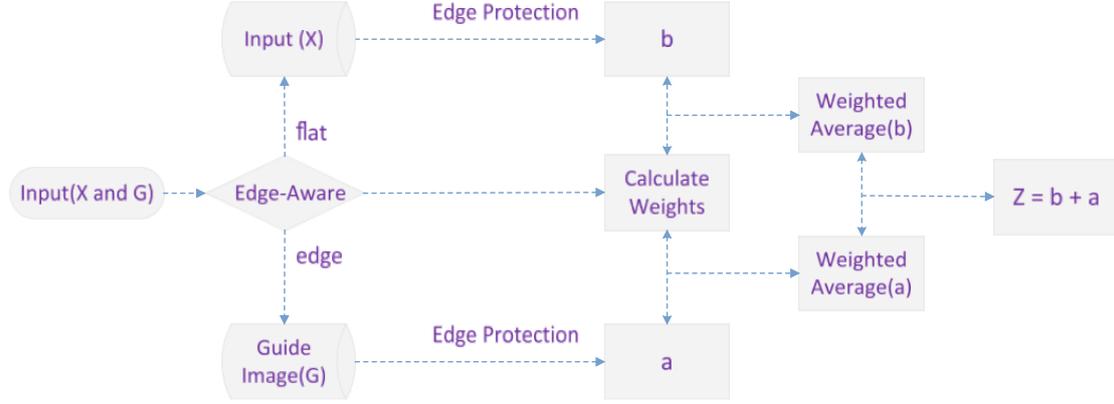

Figure. 1. Flow chart of the proposed algorithm

1) Edge-Aware Constraints

Compared with WGIF[7] and GDGIF[8], the proposed edge-aware constraint combines the coefficient of variation and gradient information to more accurately represent the fine edge information of an image. The edge perception constraint is determined by the coefficient of variation within the 3*3 window of the gradient domain image, the coefficient of variation of the input image within the $(2\xi 1+1) * (2\xi 1 +1)$ window and the gradient image. The formula is defined as follows.

$$\Gamma_G(p') = \frac{1}{N}\sum_{p=1}^{N}\frac{\zeta(p') + \varepsilon}{\zeta(p) + \varepsilon} \tag{9}$$

$$\zeta(p') = \varphi_{G,3}\varphi_{G,\xi 1}g \tag{10}$$

where $\varphi_{G,3} = \frac{\sigma_{G,3}(g_{p'})}{mean_{\sigma_{G,3}}}$ denotes the coefficient of variation of the gradient information corresponding to a window radius of 3 at pixel $p'$. $\varphi_{G,\xi 1} = \frac{\sigma_{G,\xi 1}(p')}{mean_{\sigma_{G,\xi 1}}}$ denotes the coefficient of variation at pixel $p'$ with radius $\xi 1$. $g$ denotes the gradient information obtained from the gradient calculation of the guide image, which can reflect the edge information of the image to a certain extent. The formula is shown below.

$$g_x = \frac{\partial f(x,y)}{\partial x} = f(x+1,y) - f(x,y) \tag{11}$$

$$g_y = \frac{\partial f(x,y)}{\partial y} = f(x,y+1) - f(x,y) \tag{12}$$

$$g_{(x,y)} = sqrt(g_x{}^2 + g_y{}^2) \tag{13}$$

In the context of gradient information computation, relying solely on horizontal and vertical gradient calculations may result in a considerable loss of relevant information. To accurately obtain gradient information while maintaining algorithmic efficiency, gradient calculation in all four directions is utilized.

Commonly employed for gradient computation, templates have been extensively utilized, and operators like the Robert, Prewitt, Sobel, and Canny are frequently employed. However, such gradient computation involves template calculation, which necessitates local image traversal, inevitably leading to algorithmic complexity. To address this, this paper adopts a global-based fast gradient computation approach in all four directions. Specifically, the entire image is subjected to row- and column-based difference operations, obviating the need for template operation traversal and reducing algorithmic complexity. The final gradient computation proceeds as follows.

$$g_{(x,y)} = sqrt(g_0{}^2 + g_1{}^2 + \cdots + g_3{}^2) \qquad (14)$$

where $g_{(x,y)}$ is the gradient at the pixel point (x, y), the $g_0$, $g_1$, $g_2$ and $g_3$ are the gradient information in the upper, lower, left and right directions, respectively.

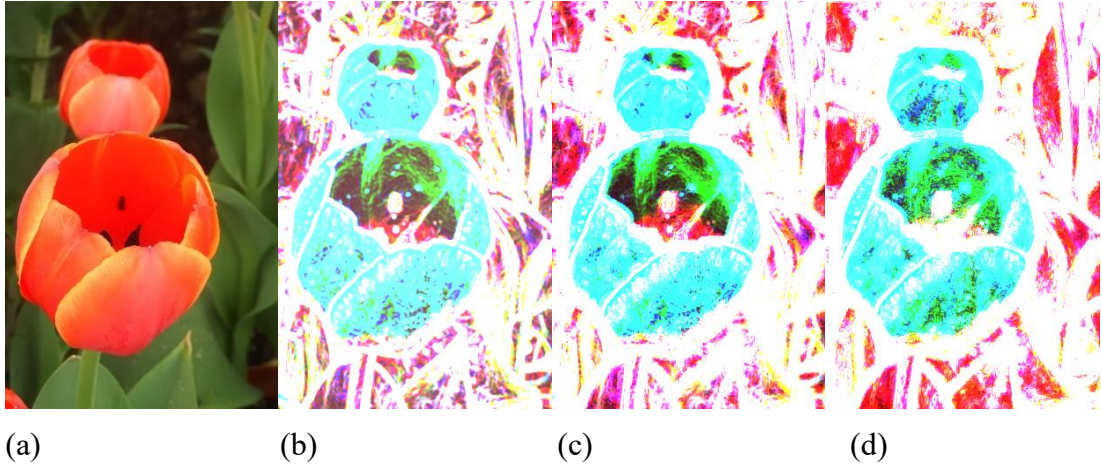

(a)      (b)      (c)      (d)

Figure. 2. Edge-aware constraint operator comparison (a) Input; (b) WGIF; (c) GDGIF; (d) Proposed

TABLE I QUANTITATIVE EVALUATION OF EDGE-AWARE OPERATORS

|  | WGIF | GDGIF | Proposed |
| --- | --- | --- | --- |
| Average gradient | 14.45 | 27.58 | 66.88 |

Fig. 2 displays a comparison of edge-aware constraint operators employed in WGIF,

GDGIF, and the algorithm described in this paper. An objective description is made using the average gradient, whereby a larger average gradient implies that an image contains more texture information, making it more effective in edge protection. Evidently, the edge-aware constraint proposed in this paper can accurately extract edge detail information from an image, as illustrated by the greater texture details within the petal in (d) compared to (b) and (c), with sharper texture edges. Table I demonstrates a significant improvement in objective evaluation metrics with the proposed edge-aware constraint operator. Consequently, when processing scenes with complex information, the proposed edge-aware constraint can better capture image detail information, enhancing sharpness while mitigating halo artifacts when processing edge regions.

2) Edge Protection Constraints

Based on the analysis in Section 2 for edge-preserving properties, combined with the edge-aware constraint operator proposed in 1), substituting the operator into equation (3) yields

$$a_{p'} = \frac{\mu_{G \odot X, \xi_1}(p') - \mu_{G, \xi_1}(p')\mu_{X, \xi_1}(p')}{\sigma^2_{G, \xi_1}(p') + \frac{\lambda}{\Gamma_G(p')}} \tag{15}$$

Let $\mu_{G \odot X, \xi_1}(p') - \mu_{G, \xi_1}(p')\mu_{X, \xi_1}(p') = \sigma^2_{GX, \xi_1}(p')$, Substitute into equation (14).

$$a_{p'} = \frac{\sigma^2_{GX, \xi_1}(p')}{\sigma^2_{G, \xi_1}(p') + \frac{\lambda}{\Gamma_G(p')}} \tag{16}$$

When the input guide image X is equal to the input image G, $\sigma^2_{G, \xi_1}(p') = \sigma^2_{GX, \xi_1}(p')$, the value of $a_{p'}$ is given by the regularization parameter $\lambda$ and $\Gamma_G(p')$. Due to the edge-awareness constraint $\Gamma_G(p')$, the effect of $\frac{\lambda}{\Gamma_G(p')}$ on $a_{p'}$ is smaller than the effect of the regularization parameter $\lambda$ on $a_{p'}$ when the pixel is in the edge region but the effect still exists, which results in $a_{p'}$ in the edge region being only close to 1 but not equal to 1. Therefore, to further reduce the effect of the regularization term parameter $\lambda$ on the edge pixels, the gap between the numerator denominators of $a_{p'}$ is reduced as much as possible by making the following changes to $a_{p'}$:

$$a_{p'} = \frac{\sigma^2_{GX,\xi1}(p') + \frac{\lambda}{\Gamma_G(p')}}{\sigma^2_{G,\xi1}(p') + \frac{\lambda}{\Gamma_G(p')}} \qquad (17)$$

at this point, the $a_{p'}$ of the pixels in the edge region is equal to 1. Since the $a_{p'}$ of the pixels in the flat region should converge to 0, a first-order edge protection constraint $\eta$ is introduced in order to distinguish the edge region from the flat region.

$$a_{p'} = \frac{\sigma^2_{GX,\xi1}(p') + \frac{\lambda}{\Gamma_G(p')}\eta_{(p')}}{\sigma^2_{G,\xi1}(p') + \frac{\lambda}{\Gamma_G(p')}} \qquad (18)$$

$\eta$ is defined as follows.

$$\begin{cases} \eta_{(x,y)} = 1 & g_{(x,y)} > 1.7 mean(g) \\ \eta_{(x,y)} = 0 & else \end{cases} \qquad (19)$$

where g is the gradient image obtained from in the edge-aware constraint, which is already computed in the edge-aware constraint. Therefore, g does not need additional computation. 1.7 is the threshold parameter that works best after experimental validation. A comparison with the conservation factor of the GDGIF algorithm is shown in the Fig.3.

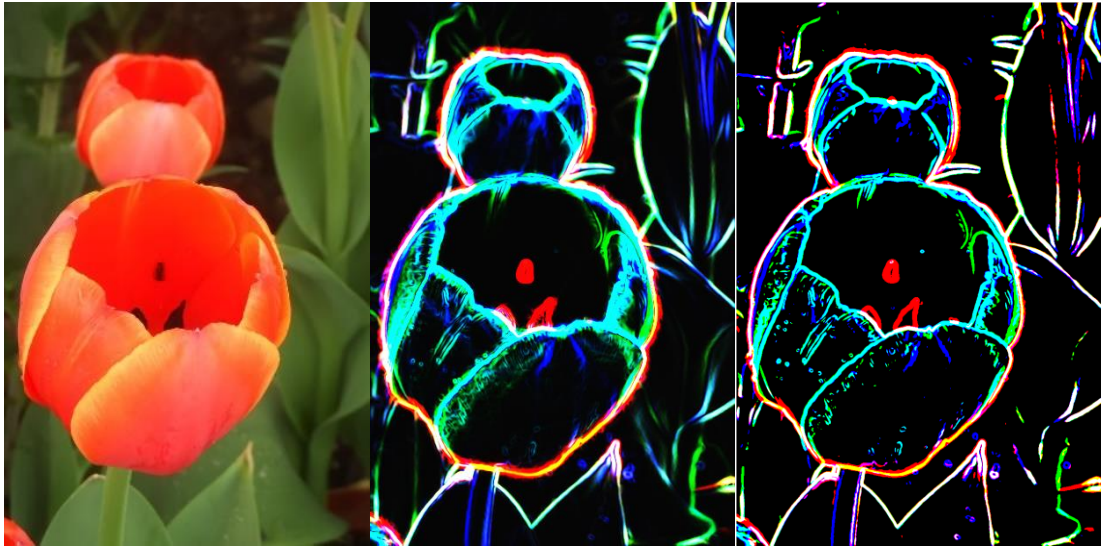

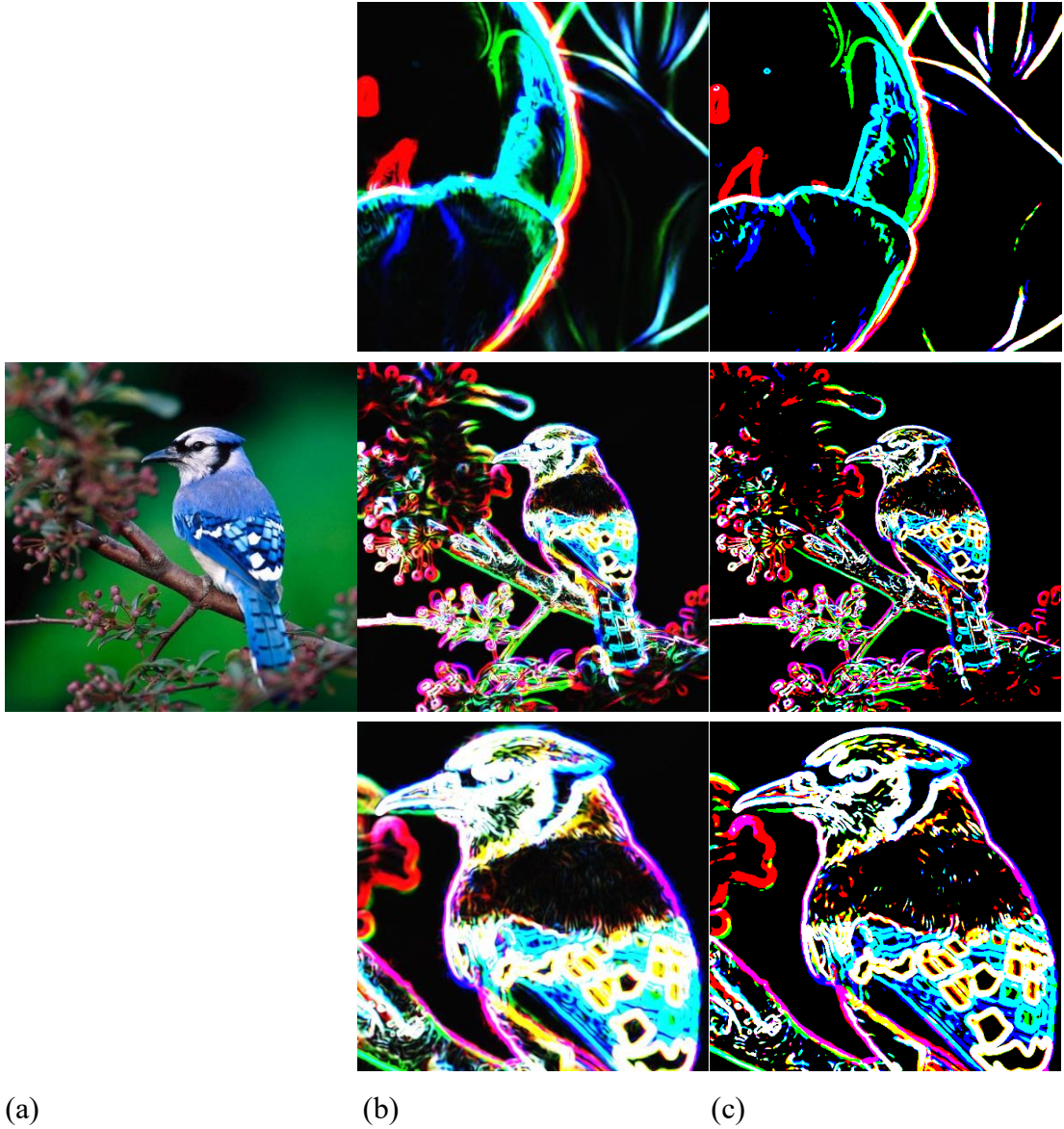

(a) (b) (c)

Figure. 3. Edge protection constraint operator comparison

TABLE II QUANTITATIVE EVALUATION OF EDGE PROTECTION OPERATORS

|  | GDGIF | Proposed |
| --- | --- | --- |
| tulips | 10.30 | 18.33 |
| bird | 16.88 | 28.01 |

Fig. 3 presents a comparison between the edge protection operator proposed in this paper and the edge protection operator in GDGIF[8]. Subjective perception and comparison of objective data demonstrate that the proposed edge protection operator extracts clearer image edges and considerably reduces edge artifacts, as illustrated by the detail enlargement in the first set of images. Objective evaluation metrics in Table

II further corroborate these observations, revealing that the proposed edge protection operator results in richer image texture information, with fewer redundant artifacts and blurred edges. In summary, the proposed edge protection operator outperforms GDGIF [8].

Combining the edge-aware operator and edge-protection operator, the final linear ridge regression model obtained by the algorithm in this paper is as follows.

$$E = \sum_{p \epsilon \Omega_{\xi_1}(p')} [\left(a_{p'}G_{(p)} + b_{p'} - X(p)\right)^2 + \frac{\lambda}{\Gamma_G(p')}(a_{p'} - \eta)^2] \quad (20)$$

3) Weighted Average Strategy

Based on the analysis of the strategy for handling overlapping windows in Section 2, a fast weighted mean strategy is proposed in this paper.

First, the template of $(2k+1) * (2k+1)$ is constructed and the absolute difference of the neighborhood sum of the center pixel in the template is calculated. Then, traverse the image and set a threshold, the image is reconstructed according to the threshold. Finally, obtain uniform and non-uniform regions, which are the weight distribution of the images. The calculation formula is as follows.

$$\mu_s = \begin{cases} 1 - \dfrac{svar_{i,j}(n)}{Th_{var}}, & svar_{i,j}(n) \leq Th_{svar} \\ thre, & svar_{i,j}(n) > Th_{svar} \end{cases} \quad (21)$$

where $svar_{i,j}(n)$ is the absolute difference of the neighborhood summation of the current center pixel. $Th_{svar}$ is the set absolute difference threshold.

When $svar_{i,j}(n)$ is less than the threshold value, it is regarded as a uniform region, and the value of this region is set to $thre$. When $svar_{i,j}(n)$ value greater than the threshold value, this region is considered as a non-uniform area.

After getting $\mu_s$ After that, it is considered as the weight distribution of the input image, and the weighted image is obtained by doing inner product with the input image, and finally, the weighted image is normalized separately to obtain the final weighted mean output with the following formula.

$$Z_{(p)} = \frac{1}{|\Omega_{\xi 1}|} \sum_{p|p\epsilon\Omega_{\xi 1}(p')} \mu_s(a_{p'}G_{(p)} + b_{p'}) \quad (22)$$

where $|\Omega_{\xi 1}| = \sum_{p|p\epsilon\Omega_{\xi 1}(p')} \mu_s$ is the normalized coefficient of the weight distribution $\mu_s$. Consequently, the equation (21) can be written as

$$Z_{(p)} = \widetilde{a_{p'}}G_{(p)} + \widetilde{b_{p'}} \quad (23)$$

where $\widetilde{a_{p'}} = \sum_{p|p\epsilon\Omega_{\xi 1}(p')} \mu_{s1} a_{p'}$, $\widetilde{b_{p'}} = \sum_{p|p\epsilon\Omega_{\xi 1}(p')} \mu_{s2} b_{p'}$. $\mu_{s1}$ and $\mu_{s2}$ are the weights of $a$ and $b$ the pixel $p'$.

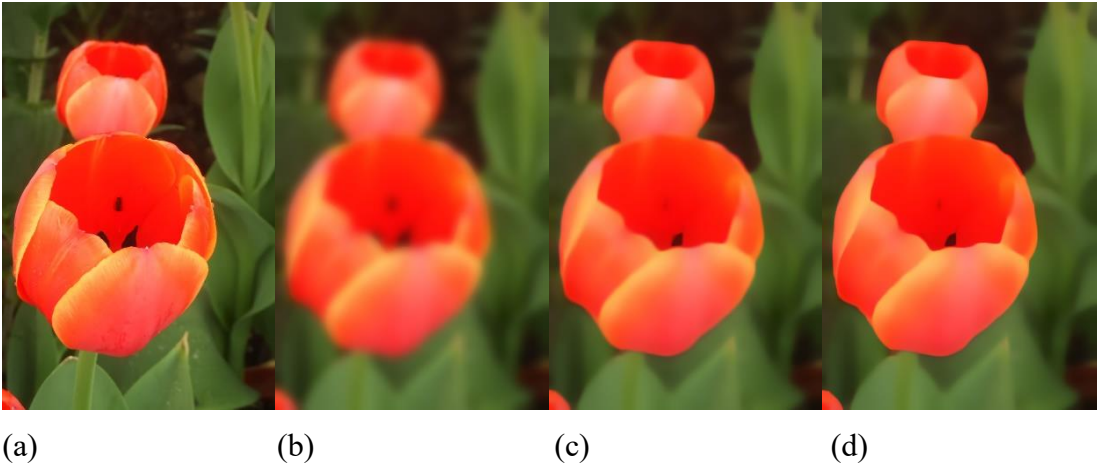

(a)      (b)      (c)      (d)

Figure. 4. Average strategies comparison $r = 16$, $\lambda = 1$, thre $= 0$, $k = 7$, $Th_{svar} = 11$, (a) input image; (b) GIF; (c) WAGIF; (d) Proposed

TABLE III QUANTITATIVE EVALUATION OF THE AVERAGE STRATEGY

|  | GIF | WAGIF | Proposed |
| --- | --- | --- | --- |
| PSNR | 18.2177 | 19.9936 | 20.0730 |
| SSIM | 0.9571 | 0.9679 | 0.9684 |

Figure 4 presents a comparison of the weighting strategies employed by GIF[6], WAGIF[10], and the proposed averaging strategy. The proposed averaging strategy utilizes the absolute difference of image neighborhood for weight distribution calculation, resulting in sharper edge distribution compared to GIF[6]. This characteristic effectively avoids the problem of excessive blurring of image edge detail information when processing with large GIF windows and further avoids the occurrence of edge halo artifacts. Objective evaluation indicators in Table III reveal that the proposed averaging strategy achieves higher structural similarity and signal-to-noise ratio compared to the WAGIF strategy. Thus, the images obtained through the proposed averaging strategy are closer to the real input images and have superior image quality.

The proposed averaging strategy employs a more accurate weighted average approach that reasonably estimates the distribution of each pixel in the window. As a result, the final image quality is significantly enhanced. Table metrics present the PSNR and structural similarity measures of this algorithm strategy and others. The objective evaluation indices demonstrate that the proposed averaging strategy significantly improves the signal-to-noise ratio and the average gradient compared to other algorithms, particularly compared to GIF. This finding affirms that the proposed strategy effectively enhances image quality and mitigates excessive blurring of image edges.

## 4. ALGORITH ANALYSIS

In summary, this section analyzes the gradient domain weighted guided filtering algorithm. Assume that the input image X is the same as the guide image G. In this case, the output image mainly depends on $a_{p'}$ in the linear model

$$a_{p'} = \frac{\sigma_{GX,\xi1}^2(p') + \frac{\lambda}{\Gamma_G(p')}\eta_{(p')}}{\sigma_{G,\xi1}^2(p') + \frac{\lambda}{\Gamma_G(p')}} \tag{24}$$

1) The Margins

The value of $\Gamma_G(p')$ is large. $\sigma_{GX,\xi1}^2 = \sigma_{G,\xi1}^2 >> \frac{\lambda}{\Gamma_G(p')}$ and $\eta_{(p')} = 1$, then

$$a_{p'} = \frac{\sigma_{GX,\xi1}^2(p') + \frac{\lambda}{\Gamma_G(p')}}{\sigma_{G,\xi1}^2(p') + \frac{\lambda}{\Gamma_G(p')}} = 1 \tag{25}$$

Due to the edge protection factor $\eta_{(p')}$, in response to a well-defined edge region, the $a_{p'}$ will be constrained to 1. Meanwhile,

$$b_{p'} = \mu_{X,\xi1}(p') - a_{p'}\mu_{G,\xi1}(p') = \mu_{X,\xi1}(p') - \mu_{G,\xi1}(p') = 0 \tag{26}$$

$$Z_{(p)} = a_{p'}G_{(p)} + b_{p'} = G_{(p)} \tag{27}$$

The output is equal to the guide image, which accurately achieves the purpose of protecting the edges. $a_{p'}$ is no longer affected by the regularization parameter $\lambda$. The effect brought by the consistency of the regularization parameters and the halo artifacts caused by GIFs using variance to judge edges is effectively avoided.

2) The Flat Area

$\Gamma_G(p')$ is close to 0, $\sigma^2_{GX,\xi 1} = \sigma^2_{G,\xi 1} \ll \frac{\lambda}{\Gamma_G(p')}$, $\eta_{(p')} = 0$, then

$$a_{p'} = \frac{\sigma^2_{GX,\xi 1}(p') + \frac{\lambda}{\Gamma_G(p')}\eta_{(p')}}{\sigma^2_{G,\xi 1}(p') + \frac{\lambda}{\Gamma_G(p')}} = \frac{\sigma^2_{GX,\xi 1}(p')}{\sigma^2_{G,\xi 1}(p') + \frac{\lambda}{\Gamma_G(p')}} \tag{28}$$

Regularization parameters $\lambda$ will affect the $a_{p'}$. However, compared to GIF、GDGIF、WGIF、and Xie, the edge perception factor proposed $\Gamma_G(p')$ are closer to the realistic flat region pixels：

In uniform flat areas, the $\Gamma_G(p')$ is smaller and the $\frac{\lambda}{\Gamma_G(p')}$ is larger. Therefore, $b_{p'} = \mu_{X,\xi 1}(p') - a_{p'}\mu_{G,\xi 1}(p') \approx \mu_{X,\xi 1}(p')$, which means that the pixels in the flat region are almost independent of the guide image $G$ and depend more on the mean value of the input image $X$ in window $\Omega_{\xi 1}$.

In the flat area near the edge, the $\Gamma_G(p')$ is larger. The value of $\frac{\lambda}{\Gamma_G(p')}$ is smaller, resulting in $a_{p'} \approx 1$. Therefore,

$$b_{p'} = \mu_{X,\xi 1}(p') - a_{p'}\mu_{G,\xi 1}(p') \approx \mu_{X,\xi 1}(p') - \mu_{G,\xi 1}(p') \approx 0 \tag{29}$$

$$Z_{(p)} = a_{p'}G_{(p)} + b_{p'} \approx G_{(p)} \tag{30}$$

This means that the output of the flat region is almost independent of the input image $X$ and depends more on the guide image $G$.

### 3) Weighted Average Strategy

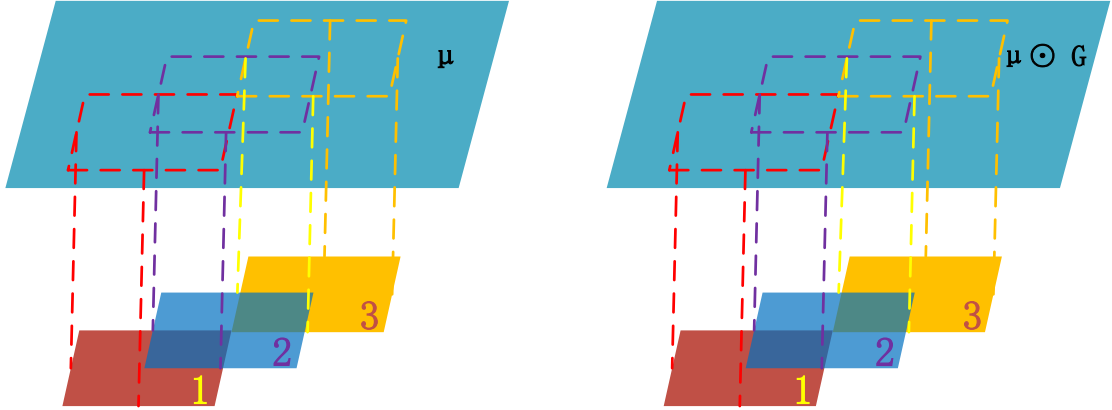

Figure. 5. Weighted Average Strategy

After calculating each pixel point's $a_{p'}$ and $b_{p'}$, the weighted averaging[28] strategy proposed is used for processing: as shown in Fig.5, G is the input guide image, $\odot$ denotes the inner product, $\mu$ denotes the weight distribution map. Fig.5 illustrates that the central pixel is concurrently calculated and assigned in different windows (1, 2, 3). The left side of the figure presents the normalization coefficients, while the right side depicts the weight output. Dividing them yields the final weighted mean output.

The proposed strategy calculates the weight distribution map through global calculation while computing pixel weights. Next, it directly derives the weighted output from the inner product of the weight distribution map and the original map, avoiding the need to traverse all image windows to assign values. Finally, the output is rapidly computed using the boxfilter presented in [6]. The complexity of this strategy is O(N) for images with N pixels, which is consistent with GIF [6] owing to the boxfilter used. Consequently, this strategy has a shorter running time and effectively enhances image quality.

### 4) Validation

Figure 6 provides one-dimensional data illustrations and objective evaluation index data of the proposed algorithm and other algorithms to verify the validity of the above analysis.

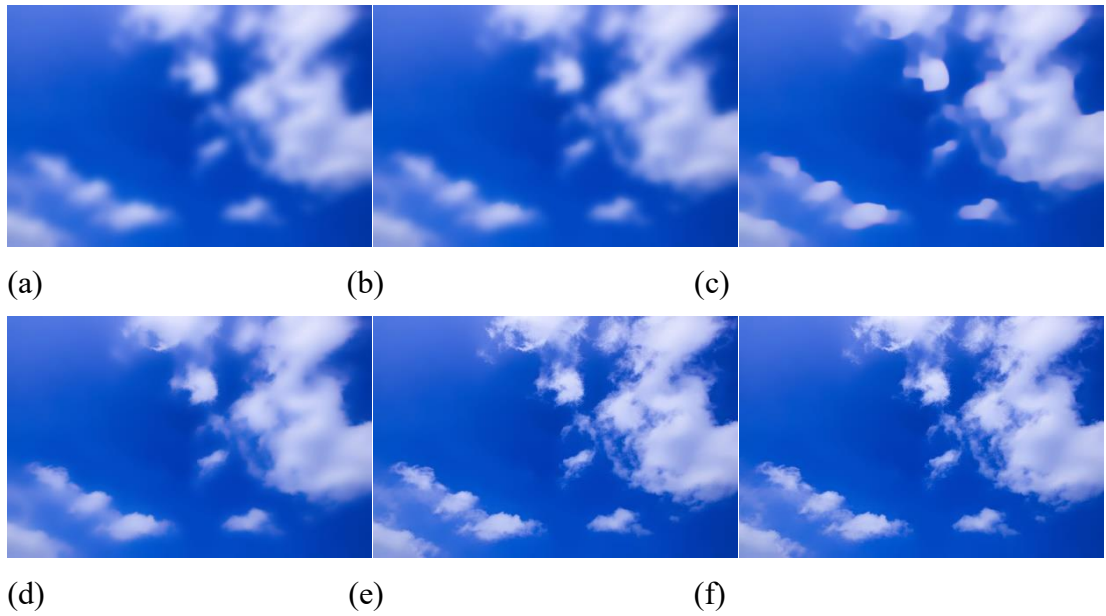

(a)  (b)  (c)

(d)  (e)  (f)

Figure. 6. Processing Effects of Different Guided Filtering Algorithms

TABLE IV QUANTITATIVE EVALUATION OF DIFFERENT EDGE-PRESERVING SMOOTHING ALGORITHMS

|      | GIF    | Xie    | WAGIF  | WGIF   | GDGIF  | Proposed |
|------|--------|--------|--------|--------|--------|----------|
| PSNR | 25.42  | 25.76  | 25.95  | 28.78  | 35.00  | 37.93    |
| SSIM | 0.9794 | 0.9807 | 0.9820 | 0.9899 | 0.9976 | 0.9982   |

$r = 16$, $\lambda = 1$  (a) GIF; (b) Xie; (c) WAGIF; (d) WGIF; (d) GDGIF; (d) Proposed;

Fig.6 shows the comparison of the processing effect between the proposed algorithm and other algorithms, and we set the parameters to be consistent for fairness ($r = 16$, $\lambda = 1$). Both subjective perception and objective data show that the proposed algorithm has sharper edges in detail regions and higher overall image quality, which can bring better visual experience. Table IV reveals that the proposed algorithm yields the highest objective evaluation index among the compared algorithms.

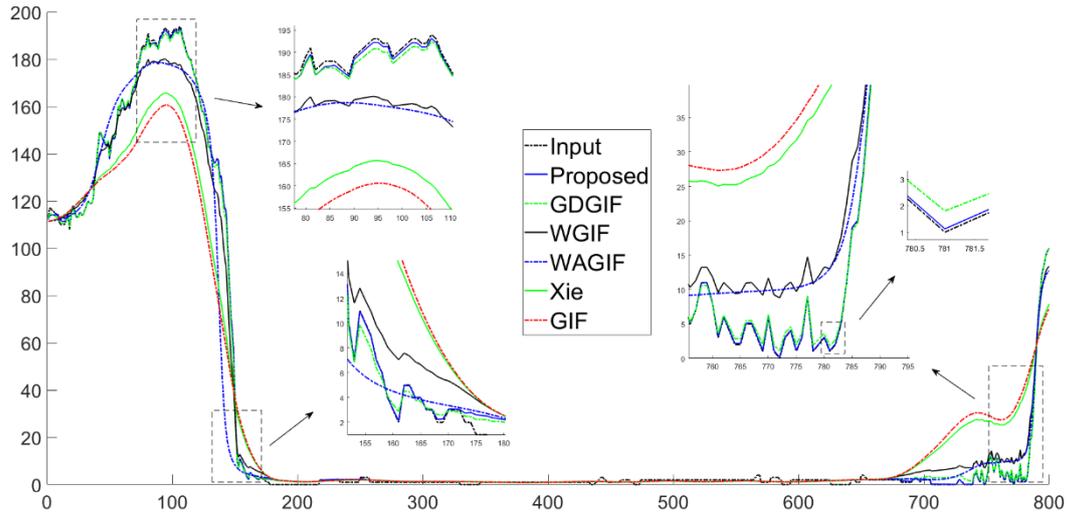

Figure. 7. Comparison of the 1D data illustrations in Fig. 6. $r = 16$, and $\lambda = 1$; The input image is derived from the red channel of the original input image in Fig. 6(a)

Fig.7 demonstrates that, in comparison to the GIF, Xie, WAGIF, GDGIF, and WGIF algorithms, the proposed algorithm more closely aligns with the original input image in the edge region, whereas the corresponding positions' pixel distribution of other algorithms is notably distant from the original input image. Further, when considering zoomed-in details, the proposed algorithm is the most aligned with the input image. In other words, the edges of the image processed by this algorithm are close to the edges of the input image. The value of $a_{p'}$ is closer to 1 in the edge region. This means that the edge of the algorithm proposed is closer to the input image and the edge protection effect of this algorithm is better. In summary, the algorithm proposed (GWGIF) can effectively maintain the edge sharpness and avoid the appearance of over-smoothing and halo artifacts, which verifies the superiority and reasonableness of this algorithm in edge protection.

## 5. APPLICATION OF THE NEW FILTER

### 1) Image Detail Enhancement

As GIF follows a linear model, the edge structure of the input image is transferred to the output image. Consequently, the difference between the input and output images can be viewed as a detail layer, which, when superimposed on the input image, can enhance image details.

According to the analysis in Section 2, the consistency of the regularization parameter $\lambda$ causes the GIF to have certain defects at the edges: the anisotropy of the pixel distribution is ignored when representing the edges, and the edge information cannot be extracted accurately. Linear scaling further amplifies the deficiency in edge information, as evidenced by black edges and edge distortion caused by excessive brightening [6]. These results will be demonstrated in later experiments.

The effect of different regularization parameters $\lambda$ on image detail enhancement is shown in Fig.8.

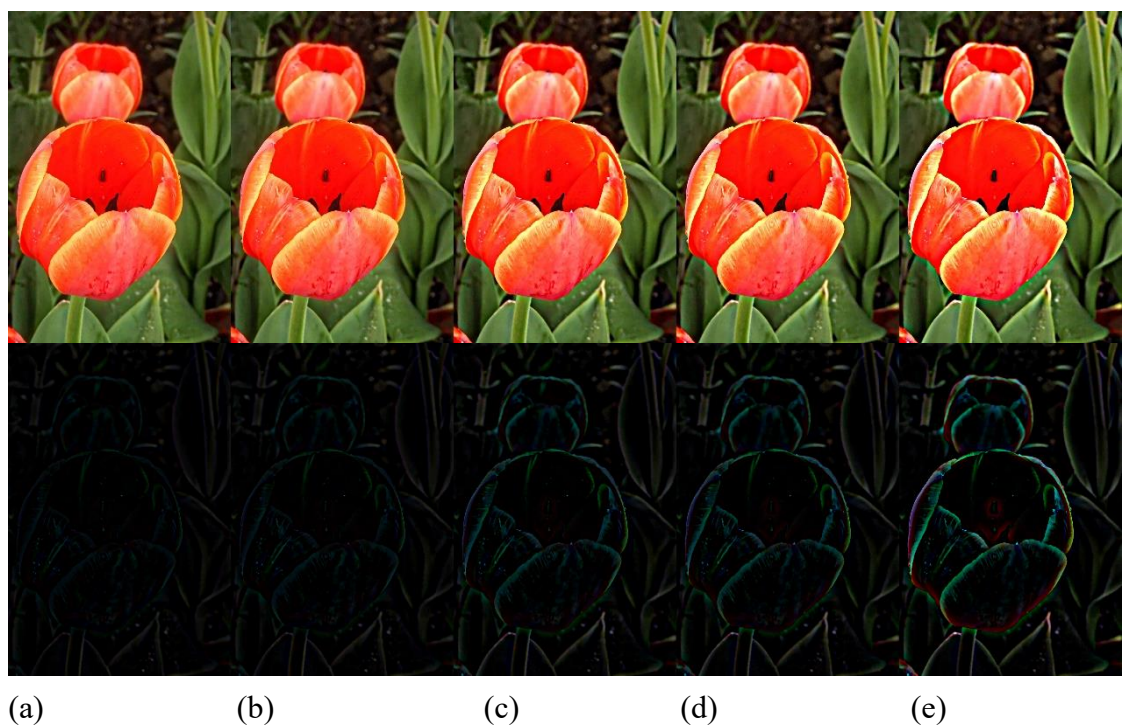

(a)　　　　　(b)　　　　　(c)　　　　　(d)　　　　　(e)

Figure. 8. Difference in guided filtering $\lambda$ Comparison  (a) $\lambda = 0.04\char`\^2$; (b) $\lambda = 0.05\char`\^2$; (c) $\lambda = 1/128$; (d) $\lambda = 0.01$; (d) $\lambda = 0.04$;

TABLE V QUANTITATIVE EVALUATION OF GUIDED FILTERING WITH DIFFERENT PARAMETERS $\lambda$

|  | 0.04^2 | 0.05^2 | 1/128 | 0.01 | 0.04 |
|---|---|---|---|---|---|
| GIF | 67.6986 | 66.8689 | 52.9433 | 47.7353 | 37.5353 |

As shown in the Fig. 8, with the increase of $\lambda$, the image presents richer edge information. However, the resulting edge artifacts are also more serious, and the edges of the image are sharpened, losing the original structural information of the image. Although perceived visual improvements may suggest that the image contrast has been enhanced, it is essential to acknowledge that image quality is a multifaceted concept that cannot be solely evaluated based on contrast. Therefore, to ensure a comprehensive and objective evaluation, we employed the Blind Image Quality Index (BIQI) metric, which has been previously utilized in studies [7, 8, 25]. A higher score on the BIQI indicates a superior image quality, and we employed this metric to compare and contrast the efficacy of different algorithms. As shown in the Table V, the BIQI score decreases as $\lambda$ increases, proving the conclusion that the larger $\lambda$ is, the lower the image quality and the more serious the halo artifacts are, which means that $\lambda$ affects the edge-preserving nature of GIF.

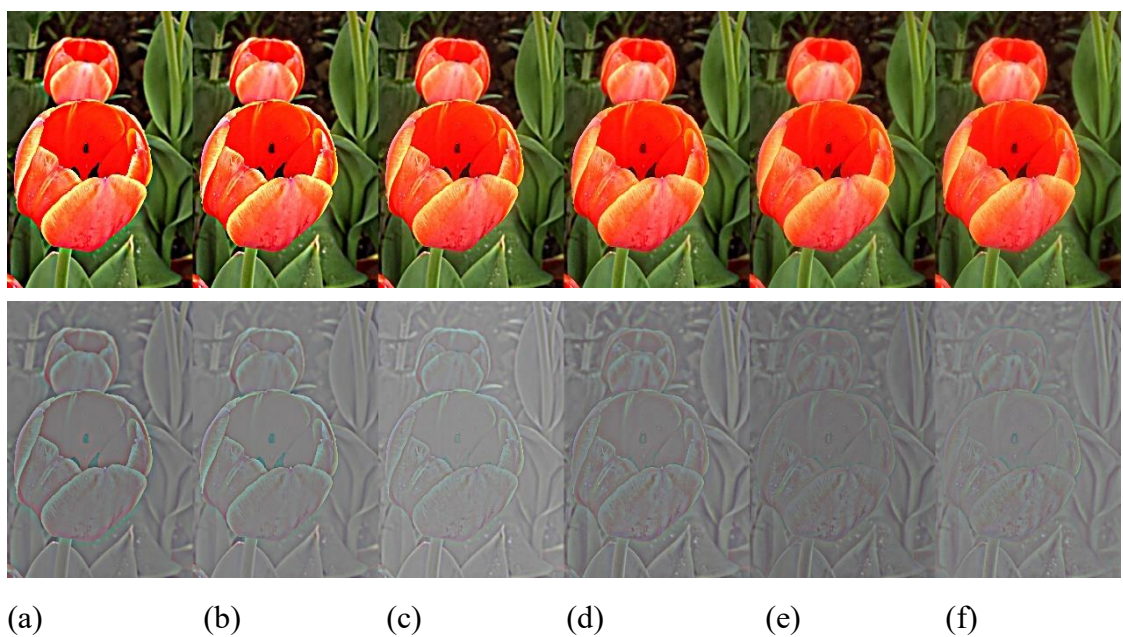

(a)      (b)      (c)      (d)      (e)      (f)

Figure. 9. Comparison of detail enhancement results, r=16. $\lambda = 0.04$. (a) GIF, (b) GDGIF, (c) WGIF, (d) Xie, (e) WAGIF, (f) Proposed

TABLE VI QUANTITATIVE EVALUATION OF DIFFERENT ALGORITHMS FOR IMAGE ENHANCEMENT

|      | GIF     | WAGIF   | Xie     | WGIF    | GDGIF   | Proposed |
|------|---------|---------|---------|---------|---------|----------|
| BIQI | 37.5353 | 41.4550 | 53.9159 | 66.4462 | 69.8050 | 71.6857  |

Figure 9 illustrates the superimposition of the detail layer into the input image, where the first row of the figure presents the comparison of different algorithms in terms of detail enhancement, while the second row shows the corresponding local enhancement of the detail layer image. Notably, despite the overall similarity in the enhancement effect of all algorithms, a greater emphasis is placed on the local edge regions, which exhibit more significant gaps. For instance, in the petal edge regions of the GIF and WAGIF images, the algorithms severely sharpened the images, resulting in a more pronounced black edge effect. In contrast, our algorithm contains fewer but sharper edge details, leading to enhanced sharpness in the edge regions. Thus, the proposed approach avoids linear enhancement that incorporates excessive information at the edge, thereby enhancing the image quality and preventing the degradation of image quality caused by sharpening and artifacts. As supported by the data presented in Table VI, our algorithm achieved the highest index score, thereby indicating that it provides the most effective image enhancement.

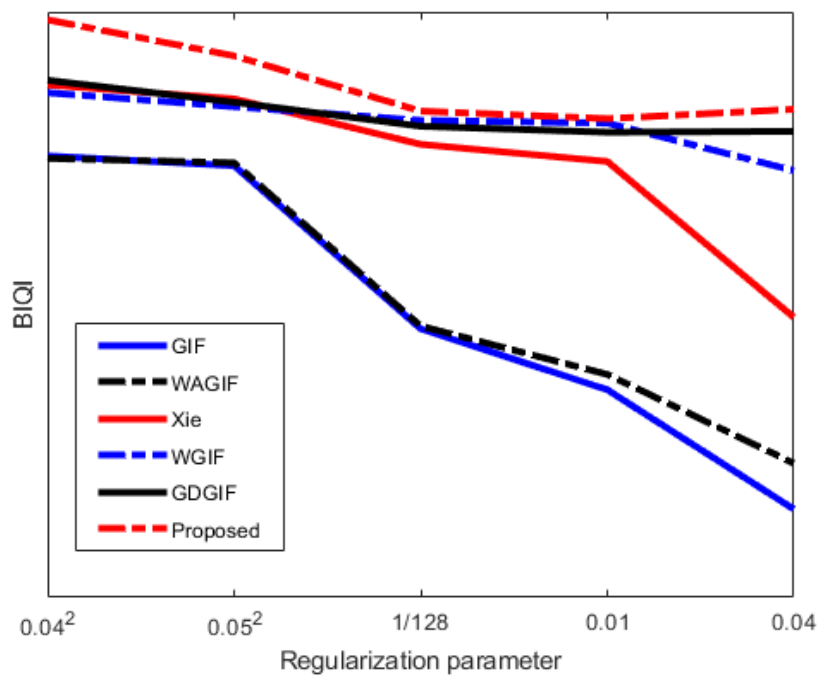

Figure. 10. Comparison of algorithm scores for different $\lambda$ parameters

TABLE VIII QUANTITATIVE EVALUATION OF IMAGE ENHANCEMENT BY DIFFERENT ALGORITHMS WITH DIFFERENT $\lambda$

|       | 0.04^2  | 0.05^2  | 1/128   | 0.01    | 0.04    |
|-------|---------|---------|---------|---------|---------|
| GIF   | 67.6986 | 66.8689 | 52.9433 | 47.7353 | 37.5353 |
| WAGIF | 67.4945 | 67.1120 | 53.1729 | 49.0460 | 41.4550 |
| Xie   | 73.7712 | 72.5683 | 68.6918 | 67.2269 | 53.9159 |
| WGIF | 73.0985 | 71.9041 | 70.7354 | 70.5101 | 66.4462 |
| GDGIF | 74.1637 | 72.2886 | 70.2363 | 69.7229 | 69.8050 |
| Proposed | 79.3267 | 76.2469 | 71.5277 | 70.8845 | 71.6857 |

According to the previous analysis: the larger the value of $\lambda$, the lower the image quality and the more serious the halo artifacts. As shown in the Fig. 10, the comparison experiment of algorithm enhancement effect with different parameters $\lambda$ is shown. As $\lambda$ increases, the image quality index (BIQI) of all algorithms in the Table VIII decreases, among which, the index of GIF algorithm decreases the most and the index of this algorithm decreases the least. It is proved that: as the regularization parameter $\lambda$ increases, the algorithm proposed in this paper is less affected and has obvious optimization of the regularization parameter $\lambda$. According to the previous analysis: the edge artifact phenomenon of the guided filtering algorithm mainly comes from the consistency of the regularization parameter $\lambda$. In summary, the proposed algorithm makes the edge artifact phenomenon better suppressed and outperforms the GIF in [6], the WAGIF in [10], Xie[9], the GDGIF in [8] and the WGIF in [7] algorithms in image detail enhancement.

2) Image Denoising

Given that the GIF algorithm exhibits edge-preserving properties, it has the ability to eliminate noise while preserving edge information. To assess the effectiveness of the proposed algorithm, we performed a grayscale image denoising experiment using the grayscale image of Lena.

To be fair, we set the experimental parameters to be consistent (r = 9, $\lambda = 0.1$ ), and superimposed a Gaussian noise with a standard deviation of 25 on the experimental images.

Fig.11 demonstrates that the proposed algorithm achieved the most optimal output when tested on noisy images, and the objective evaluation metrics, including peak signal-to-noise ratio and structural similarity, also yielded the best results. This indicates that, compared to other algorithms, our proposed approach can effectively remove noise while preserving image quality.

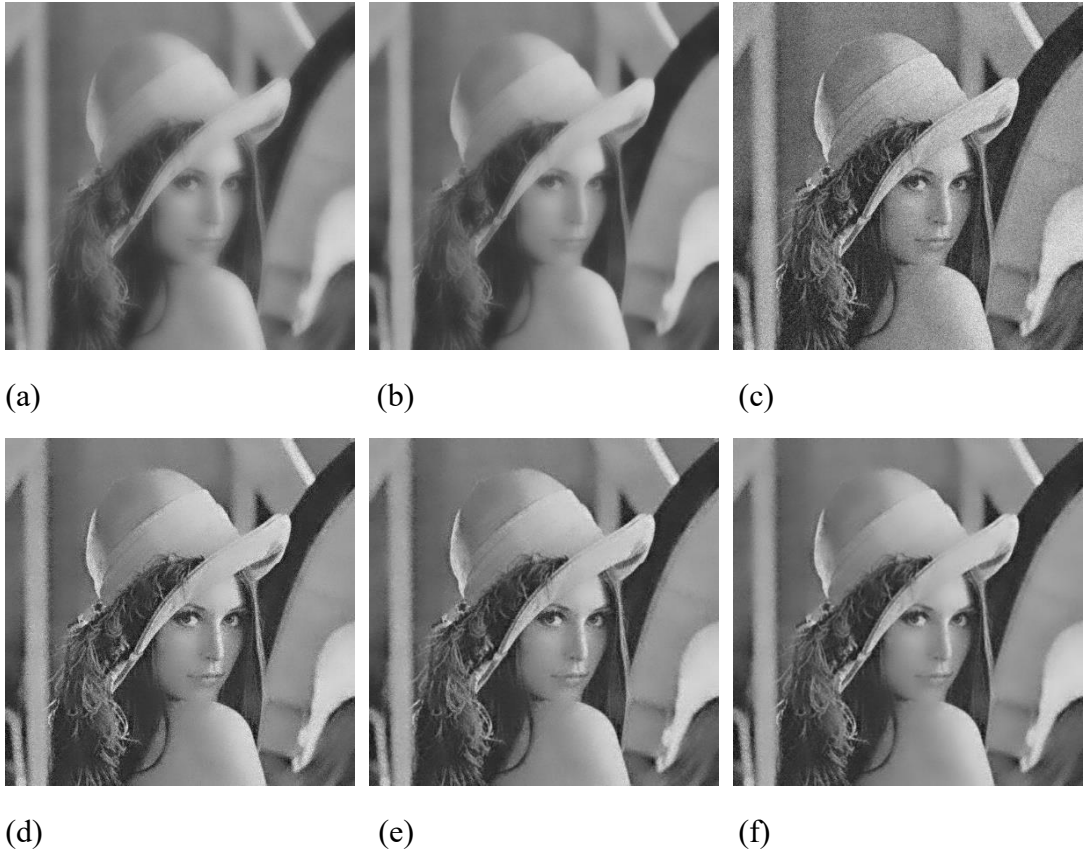

(a) (b) (c)

(d) (e) (f)

Figure. 11. Denoising effect of different algorithms. r = 9, $\lambda = 0.1$ (a) GIF (b) WAGIF (c) Xie (d) GDGIF (e) WAGIF (f) Proposed

TABLE IX QUANTITATIVE EVALUATION OF DIFFERENT ALGORITHMS FOR IMAGE DENOISING

|      | NOISE  | Xie    | WAGIF  | GIF    | GDGIF  | WGIF   | Proposed |
|------|--------|--------|--------|--------|--------|--------|----------|
| PSNR | 18.86  | 20.67  | 22.10  | 22.20  | 22.34  | 22.43  | 22.65    |
| SSIM | 0.8731 | 0.9193 | 0.9475 | 0.9447 | 0.9434 | 0.9456 | 0.9481   |

In processing noisy images, the application of the mean value strategy in GIF often leads to the generation of excessively blurred images, as depicted in Fig.11(a). By incorporating the fast weighted averaging strategy and the edge retention operator, the proposed algorithm demonstrated remarkable preservation of pixels in edge regions, as evidenced in Fig.11. Notably, the algorithm significantly enhances imaging quality in flat regions. The effectiveness of the proposed approach is further corroborated by the superior performance revealed by objective evaluation indices presented in Table IX.

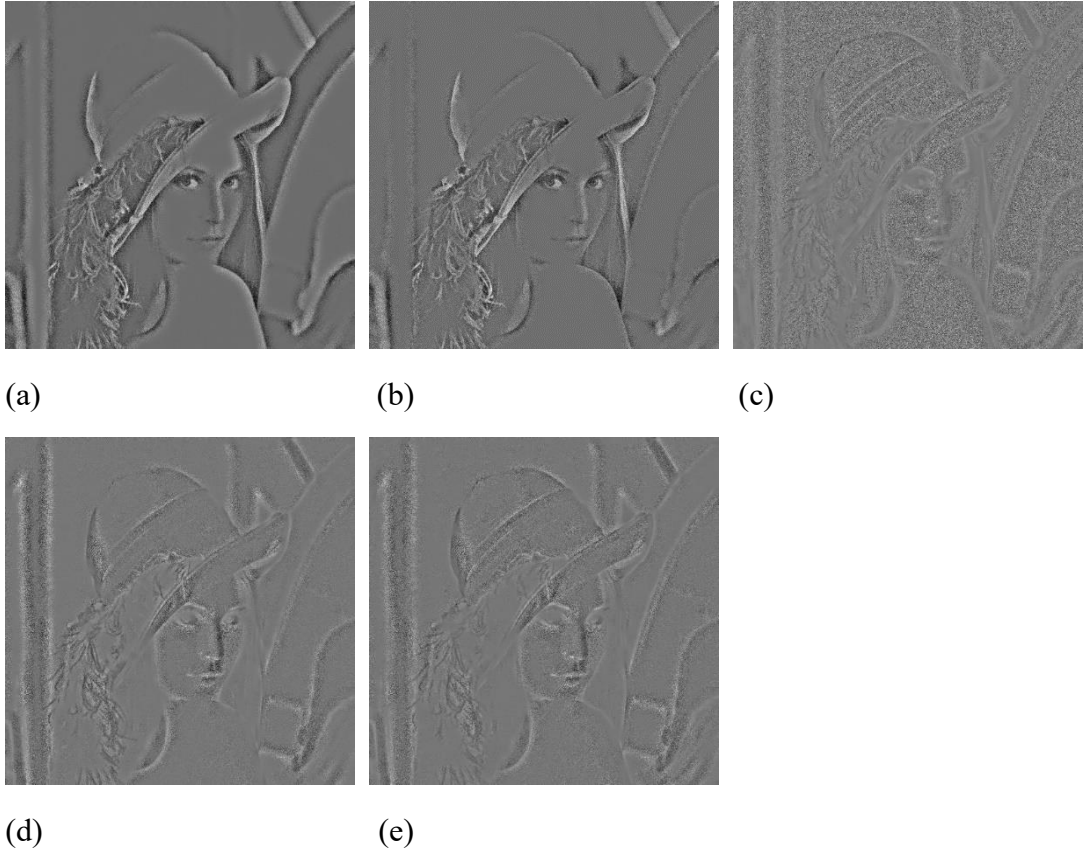

Figure. 12. Difference diagram of the proposed algorithm and other edge-preserving algorithms. r = 9, $\lambda = 0.1$ (a) GIF (b) WAGIF (c) Xie (d) GDGIF(e) Proposed

To further demonstrate the edge-preserving denoising capability of the proposed algorithm, Fig.12 presents a comparison of the difference plots between this algorithm and other edge-preserving methods. The difference plots provide a visualization of the dissimilarities in the results between the proposed algorithm and other methods. A more detailed image representation in the difference plots indicates superior preservation of image details by the proposed algorithm, thereby underscoring its remarkable edge-preserving characteristics.

In Fig.12, a comparison of the Lina contour information between the proposed algorithm and other denoising methods is presented. Fig.12 (a) and Fig.12 (b) show that the Lina contour information is more apparent in the proposed algorithm due to its superior edge preservation capability. Conversely, the poor edge-preserving performance of another algorithm leads to the erasure of edge information. In Fig.12 (c), noise is still present, indicating ineffective noise removal. Fig.12 (d) and Figure 12 (e) reveal partial edge information, which underscores the improved edge preservation of the proposed algorithm over the WGIF and GDGIF algorithms. Collectively, the results demonstrate that the proposed algorithm surpasses the performance of the GIF algorithm in [6], WAGIF algorithm in [10], Xie algorithm in [9], GDGIF algorithm in [8], and the WGIF algorithm in [7] for image denoising.

## 6. CONCLUSION

To address the limitation of the guided filtering algorithm in preserving edges, this study proposes a novel approach called GWGIF. GWGIF is a guided filtering algorithm that combines gradient and weight information in the edge constraint to accurately preserve the sharp edges of the image. Unlike previous guided filters, such as GDGIF and WGIF, which only introduce either gradient or weight information, the proposed approach incorporates both, leading to more precise edge preservation. Moreover, to enhance the accuracy of edge detection, we propose a new image weight assignment strategy and edge constraint operator. The proposed weight assignment strategy enables accurate identification of edge regions in the image, while the edge constraint operator helps to smooth the pixels in flat regions reasonably. Our experiments on image detail enhancement and image noise removal demonstrate that the images generated by the GWGIF filter have better visual perception and image quality than those produced by other guided filter-based algorithms. Given the superior performance of guided filters, we anticipate that the GWGIF algorithm can be applied to a wide range of fields, such as image defogging [26],fusion of differently exposed images, joint upsampling[27], and tone mapping of HDR images[28],among others. In the future, we will continue to investigate related issues to further enhance the accuracy and effectiveness of the proposed method in various image processing applications.

**Funding**

This work was supported by National Natural Science Foundation of China (Grant no. 62105152)，Fundamental Research Funds for the Central Universities (Grant no. 30919011401,30922010204， 30922010718， JSGP202202)，Funds of the Key Laboratory of National Defense Science and Technology (Grant no.6142113210205)，

Leading Technology of Jiangsu Basic Research Plan (Grant no.BK20192003)，The Open Foundation of Key Lab of Optic-Electronic and Communication of Jiangxi Province (Grant no.20212OEC002).

## References

[1] Zeev Farbman, Raanan Fattal, Dani Lischinski, and Richard Szeliski. 2008. Edge-preserving decompositions for multi-scale tone and detail manipulation. ACM Trans.Graph.27,3(August2008),1–10. https://doi.org/10.1145/1360612.1360666

[2] Li Xu, Cewu Lu, Yi Xu, and Jiaya Jia. 2011. Image smoothing via L0 gradient minimization. In Proceedings of the 2011 SIGGRAPH Asia Conference (SA '11). Association for Computing Machinery, New York, NY, USA, Article 174, 1–12. https://doi.org/10.1145/2024156.2024208

[3] Min, D., Choi, S., Lu, J., Ham, B., Sohn, K., Do, M.N.: Fast global image smoothing based on weighted least squares. IEEE Trans. Image Process. 23(12), 5638–5653 (2014). https://doi.org/10.1109/TIP.2014.2366600

[4] Michailovich, O.V.: An iterative shrinkage approach to total- variation image restoration. IEEE Trans. Image Process. 20(5), 1281–1299 (2011). https://doi.org/10.1109/TIP.2010.2090532

[5] Sylvain Paris, Pierre Kornprobst, Jack Tumblin and Frédo Durand (2009), "Bilateral Filtering: Theory and Applications", Foundations and Trends® in Computer Graphics and Vision: Vol. 4: No. 1, pp 1-73. http://dx.doi.org/10.1561/0600000020

[6] He, K., Sun, J., Tang, X.: Guided image filtering. IEEE Trans. Pattern Anal. Mach. Intell. 35(6), 1397–1409 (2013). https://doi.org/10.1109/TPAMI.2012.213

[7] Li, Z., Zheng, J., Zhu, Z., Yao, W., Wu, S.: Weighted guided image filtering. IEEETrans.ImageProcess.24(1),120–129(2015). https://doi.org/10.1109/TIP.2014.2371234

[8] Kou, F., Chen, W., Wen, C., Li, Z.: Gradient domain guided image filtering. IEEE Trans.ImageProcess.24(11),4528–4539(2015). https://doi.org/10.1109/TIP.2015.2468183

[9] Xie W, Zhou Y Q, You M. Improved guided image filtering integrated with gradient information[J]. Journal of Image and Graphics, 2016, 21(9): 1119-1126.

[10] Chen, B., Wu, S. Weighted aggregation for guided image filtering. SIViP 14, 491–498 (2020). https://doi.org/10.1007/s11760-019-01579-1

[11] Z. Sun，B. Han，J. Li，J. Zhang 和 X. Gao，"Weighted Guided Image Filtering with Steering Kernel，" IEEE Transactions on Image Processing， vol. 29， pp. 500-508，2020; https://doi.org/10.1109/TIP.2019.2928631

[12] Li Xu, Qiong Yan, Yang Xia, and Jiaya Jia. 2012. Structure extraction from texture via relative total variation. ACM Trans. Graph. 31, 6, Article 139 (November 2012), 10 pages. https://doi.org/10.1145/2366145.2366158

[13] D. Min, S. Choi, J. Lu, B. Ham, K. Sohn and M. N. Do, "Fast Global Image Smoothing Based on Weighted Least Squares," in IEEE Transactions on Image Processing, vol. 23, no. 12, pp. 5638-5653, Dec. 2014, A. K. Moorthy and A. C. Bovik, "A Two-Step Framework for Constructing Blind Image Quality Indices," in IEEE Signal Processing Letters, vol. 17, no. 5, pp. 513-516, May 2010; https://doi.org/10.1109/LSP.2010.2043888.

[14] Ben Weiss. 2006. Fast median and bilateral filtering. In ACM SIGGRAPH 2006 Papers (SIGGRAPH '06). Association for Computing Machinery, New York, NY, USA, 519–526. https://doi.org/10.1145/1179352.1141918.

[15] Frédo Durand and Julie Dorsey. 2002. Fast bilateral filtering for the display of high-dynamic-range images. In Proceedings of the 29th annual conference on Computer graphics and interactive techniques (SIGGRAPH '02). Association for Computing Machinery, New York, NY, USA, 257–266. https://doi.org/10.1145/566570.566574

[16] Z. Farbman, R. Fattal, D. Lischinski, and R. Szeliski, "Edge-preserving decompositions for multi-scale tone and detail manipulation," ACM Trans. Grap., vol. 27, no. 3, pp. 249–256, Aug. 2008.

[17] Pravin Bhat, C. Lawrence Zitnick, Michael Cohen, and Brian Curless. 2010. GradientShop: A gradient-domain optimization framework for image and video filtering. ACM Trans. Graph. 29, 2, Article 10 (March 2010), 14 pages. https://doi.org/10.1145/1731047.1731048

[18] P. Bhat, C. L. Zitnick, M. Cohen, and B. Curless, "GradientShop: A gradient-domain optimization framework for image and video filter-ing," ACM Trans. Graph., vol. 29, no. 2, Apr. 2010, Art. ID 10.

[19] A. Torralba and W. T. Freeman, "Properties and applications of shape recipes," in Proc. IEEE Comput. Vis. Pattern Recognit. (CVPR), Jun. 2003, pp. II-383–II-390. http://hdl.handle.net/1721.1/6695

[20] A. Levin, D. Lischinski and Y. Weiss, "A Closed-Form Solution to Natural Image Matting," in IEEE Transactions on Pattern Analysis and Machine Intelligence, vol.30,no.2,pp.228-242,Feb.2008, https://doi.org/10.1109/TPAMI.2007.1177.

[21] Ren, W., Ma, L., Zhang, J., Pan, J., Cao, X., Liu, W., Yang,M.: Gated fusion network for single image dehazing. In: 2018 IEEE Conference on Computer Vision and PatternRecognition,pp.3253–3261(2018). https://doi.org/10.1109/CVPR.2018.00343

[22] Pizarro, L., Mrázek, P., Didas, S. et al. Generalised Nonlocal Image Smoothing. Int J Comput Vis 90, 62–87 (2010). https://doi.org/10.1007/s11263-010-0337-7

[23] S. Rakshit, A. Ghosh, B.Uma Shankar, Fast mean filtering technique (FMFT),Pattern Recognition,Volume 40, Issue 3,2007,Pages 890-897, ,ISSN 0031-3203; https://doi.org/10.1016/j.patcog.2006.02.008.

[24] Matthias Kirchner and Jessica Fridrich "On detection of median filtering in digital images", Proc. SPIE 7541, Media Forensics and Security II, 754110 (27 January 2010); https://doi.org/10.1117/12.839100

[25] A. K. Moorthy and A. C. Bovik, "A Two-Step Framework for Constructing Blind Image Quality Indices," in IEEE Signal Processing Letters, vol. 17, no. 5, pp. 513-516, May 2010; https://doi.org/10.1109/LSP.2010.2043888.

[26] K. He, J. Sun and X. Tang, "Single Image Haze Removal Using Dark Channel Prior," in IEEE Transactions on Pattern Analysis and Machine Intelligence, vol. 33, no. 12, pp. 2341-2353, Dec. 2011, doi: 10.1109/TPAMI.2010.168.

[27] Johannes Kopf, Michael F. Cohen, Dani Lischinski, and Matt Uyttendaele. 2007. Joint bilateral upsampling. ACM Trans. Graph. 26, 3 (July 2007), 96–es. https://doi.org/10.1145/1276377.1276497

[28] Wang, Zhou and Qiang Li. "Information Content Weighting for Perceptual Image Quality Assessment." IEEE Transactions on Image Processing 20 (2011): 1185-1198.

[29] Eduardo S. L. Gastal and Manuel M. Oliveira. 2011. Domain transform for edge-aware image and video processing. ACM Trans. Graph. 30, 4, Article 69 (July 2011), 12 pages. https://doi.org/10.1145/2010324.1964964

[30] A. Levin, D. Lischinski and Y. Weiss, "A Closed-Form Solution to Natural Image Matting," in IEEE Transactions on Pattern Analysis and Machine Intelligence, vol. 30, no. 2, pp. 228-242, Feb. 2008, doi: 10.1109/TPAMI.2007.1177.

[31] R.C. Gonzalez and R.E. Woods, Digital Image Processing, second ed. Prentice Hall, 2002.

[32] Michael Kass and Justin Solomon. 2010. Smoothed local histogram filters. In ACM SIGGRAPH 2010 papers (SIGGRAPH '10). Association for Computing Machinery, New York, NY, USA, Article 100, 1–10. https://doi.org/10.1145/1833349.1778837